BEST PRACTICES IN MEDICAL IMAGING

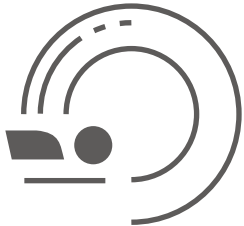

# CLINICAL ACCEPTANCE OF SOFTWARE BASED ON ARTIFICIAL INTELLIGENCE TECHNOLOGIES (RADIOLOGY)

Moscow
2019

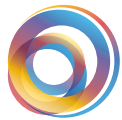

RADIOLOGY MOSCOW
ENTERPRISE IMAGING

**Developer:**
Research and Practical Clinical Center for Diagnostics and Telemedicine Technologies, Moscow Health Care Department

**Authors:**
**S.P. Morozov** – MD, MPH, PhD, Professor, CEO of the Research and Practical Clinical Center for Diagnostics and Telemedicine Technologies, Moscow Health Care Department, Chief Regional Radiology and Instrumental Diagnostics Officer, Moscow Health Care Department, Chief Regional Radiology and Instrumental Diagnostics Officer in the Central Federal District of the Russian Federation, Ministry of Health of the Russian Federation
**A.V. Vladzymyrskyy** – MD, PhD, Deputy Director for Science of the Research and Practical Clinical Center for Diagnostics and Telemedicine Technologies, Moscow Health Care Department
**V.G. Klyashtornyy** – PhD, Analyst at the Department of Scientific Activity Coordination, Research and Practical Clinical Center for Diagnostics and Telemedicine Technologies, Moscow Health Care Department
**A.E. Andreychenko** – PhD, Senior Researcher at the Department of Technical Monitoring and Quality Assurance Development, Research and Practical Clinical Center for Diagnostics and Telemedicine Technologies, Department of Health Care of Moscow
**N.S. Kulberg** – PhD, Head of the Department of Medical Imaging Tools Development, Research and Practical Clinical Center for Diagnostics and Telemedicine Technologies, Moscow Health Care Department
**V.A. Gombolevsky** – MD, PhD, Head of the Department of Radiology Quality Development, Research and Practical Clinical Center for Diagnostics and Telemedicine Technologies, Moscow Health Care Department
**K.A. Sergunova** – PhD, Head of the Department of Technical Monitoring and Control Tools Development, Research and Practical Clinical Center for Diagnostics and Telemedicine Technologies, Moscow Health Care Department

**Reviewers:**
**A.V. Gusev** – PhD, expert at Complex Medical Information Systems; Supervisory Board member at the National Database of Medical Knowledge Association of Developers and Users of Artificial Intelligence for Medicine; member of the Expert Council on the Use of Information and Communication Technology in the Healthcare System, Ministry of Health of the Russian Federation
**G.S. Lebedev** – EngD, Head of the Department of Information and Web Technologies, I.M. Sechenov First Moscow Medical State University; Advisor to the Director, State Budgetary Institution Central Research Institute for Organization and Informatization of Healthcare, Ministry of Health of the Russian Federation
**M.G. Belyaev** – PhD, Head of Data Analysis in Neuroscience, A.A. Kharkevich Institute for Information Transmission Problems, Russian Academy of Sciences
**V.A. Kutichev** – Head of the Laboratory for Medical Device Software Testing, State Budgetary Institution All-Russian Scientific Research Institute of Medical Equipment of Roszdravnadzor

**Aim:** provide a methodological framework for the process of clinical tests, clinical acceptance, and scientific assessment of algorithms and software based on the artificial intelligence (AI) technologies. Clinical tests are considered as a preparation stage for the software registration as a medical product. The authors propose approaches to evaluate accuracy and efficiency of the AI algorithms for radiology.

For correspondence: info@npcmr.ru, npcmr@zdrav.mos.ru
28/1, Srednyaya Kalitnikovskaya st., Moscow, 109029, Russia
+7 (495) 276-04-36

RADIOLOGY MOSCOW

# **CONTENTS**

# NORMATIVE REFERENCES

The document has references to the following regulatory documents (standards):

1. Federal Law of November 21, 2011 No. 323-FZ "On the Fundamentals of Health Protection in the Russian Federation".

2. Federal Law of July 27, 2006 No. 152-FZ "On Personal Data".

3. Federal Law of July 27, 2006 No. 149-FZ "On Information, Digital Technologies and Data Protection".

4. Federal Law of December 27, 2002 No. 184-FZ "On Technical Regulation".

5. Decree of the President of the Russian Federation of October 10, 2019 No. 490 "On the Development of Artificial Intelligence in the Russian Federation" and the National Strategy for the Artificial Intelligence Development for the period up to 2030.

6. Decree of the President of the Russian Federation of May 9, 2017 No. 203 "On the Strategy for the Information Society Development in the Russian Federation for 2017–2030".

7. Regulation of the Government of the Russian Federation of December 27, 2012 No. 1416 "On the Approval of Rules for the State Registration of Medical Products".

8. Regulation of the Government of the Russian Federation of May 5, 2018 No. 555 "On the Unified State Information System in the Field of Health Care".

9. Order of the Ministry of Health Care of the Russian Federation of January 9, 2014 No. 2n "On the Approval of Assessing the Conformity of Medical Devices in the Form of Technical Tests, Toxicological Studies, Clinical Tests for the State Registration of Medical Devices".

10. Order of the Ministry of Health Care of the Russian Federation of April 1, 2016 No. 200n "On the Approval of Rules of Good Clinical Practice".

11. Order of the Moscow Health Care Department of December 25, 2017 No. 918 "On the Regulations for Data Registration in the Unified Radiological Information Service in Medical Centers of the State Health System in Moscow".

12. GOST R ISO 14155-2014 "Clinical Studies. Good Clinical Practice" (Order of Rostekhregulirovanie of June 4, 2014 No. 497-st).

13. Letter of the Federal Service for Supervision in Health Care of December 30, 2015 No. 01I-2358/15 "On the Software Registration".

14. Methodical recommendations on the order of assessing the quality, efficiency and safety of medical devices (in terms of software) for the state registration under the national system / M.: State Budgetary Institution All-Russian Scientific Research Institute of Medical Equipment of Roszdravnadzor, 2018. – 31 p.

# GLOSSARY

The document contains the following terms with appropriate definitions:

**Artificial intelligence (AI)** refers to systems that display intelligent behavior by analyzing their environment and taking actions – with some degree of autonomy – to achieve specific goals. AI-based systems can be purely software-based, acting in the virtual world (e.g., voice assistants, image analysis software, search engines, speech and face recognition systems) or AI can be embedded in hardware devices (e.g., advanced robots, autonomous cars, drones, or Internet of Things applications) [6][1].

**Mathematical model** is an abstract mathematical representation of a process, device or concept; it uses a number of variables to represent inputs, outputs and internal states, and sets of equations and inequalities to describe their interaction.

**Machine learning** 1. A field in computer science that builds computational models that have the ability of "learning" from the data and then provide predictions. Depending on whether there is a supervisory signal, machine learning can be divided into three categories: the supervised learning, unsupervised learning, and reinforcement learning. 2. A technology for automatic learning in recognition and classification with test datasets to improve pattern detection, processing, and forecasting.

**Dataset (reference data, labeled dataset)** is a set of data that has been pre-prepared (processed) as per the legislation of the Russian Federation on information, information technologies, and information protection, and is necessary for the development of a software based on artificial intelligence[2]. In medicine and health care, it is a structured set of diagnostic data, including diagnostic images and information on pathological changes on images; structured cases of medical care and related electronic medical documents from electronic medical records; libraries of keywords, phrases and their critical combinations, data of genetic tests, as well as combinations of various medical data, combined into a depersonalized patient (the so-called "digital twin"). If the dataset contains confirmed information on the final diagnosis for each case or confirmed pathological process, then it is called "verified".

---

[1] National Strategy for the Artificial Intelligence Development for the period up to 2030 (approved by the Decree of the President of the Russian Federation of October 10, 2019 No. 490 "On the Development of Artificial Intelligence in the Russian Federation"

[2] Ibid

**Activation threshold ("cut-off" point)** is a predefined value, which triggers the algorithm (for example, the presence of a pathology).

**Labelling / tagging** is a stage of processing structured and unstructured data, during which data (including text documents, photo and video images) are assigned identifiers that reflect the data type (data classification) and/or data is interpreted to solve a specific problem, including using machine learning methods[3]. In medicine and health care, it is a standardized process of recording the pathological changes on diagnostic images to a database; the process includes lesion localization and nature, pathology reports, data extracted from unstructured text, and other information; several qualified medical specialists independently perform the process for each image or study.

**Artificial intelligence technologies (intellectual technologies)** are technologies created based on artificial intelligence, including computer vision, natural language processing, speech recognition and synthesis, intelligent decision support, and advanced artificial intelligence methods[4].

---

[3] Ibid.
[4] Ibid.

# SYMBOLS AND ABBREVIATIONS

**AI** – Artificial intelligence
**AUC** – Area Under Curve
**CAMI** – Centralized Archive of Medical Images
**CE** – Conformité Européenne
**CONSORT** – Consolidated Standards of Reporting Trials
**FDA** – Food and Drug Administration
**IT** – Information technologies
**MO** – Medical organization
**PACS** – Picture Archiving and Communication System
**PCTT** – Preliminary clinical and technical tests
**RD** – Radiology department
**RF** – Russian Federation
**RIS** – Radiology Information System
**ROC** – Receiver Operating Characteristic
**SO** – Software
**URIS** – Unified Radiological Information Service

# INTRODUCTION

In accordance with the National Strategy for the Artificial Intelligence Development for the period up to 2030 (approved by the Decree of the President of the Russian Federation of October 10, 2019 No. 490 "On the Development of Artificial Intelligence in the Russian Federation"), the use of artificial intelligence technologies in the social sphere contributes to the creation of conditions for improving the standard of living of the population, including improving the quality of health care services (preventive examinations, image-based diagnostics, disease incidence and development prediction models, selection of optimal medicine dosages, reducing the risk of pandemics, as well as automation and higher accuracy of surgical interventions).

Artificial Intelligence (AI) is a research field that focuses on a hardware or software modeling of human activities traditionally considered to be intellectual. AI is a part of computer sciences. New technologies are related to information technologies. The concept of artificial intelligence, however, is not novel. An automated analysis of medical information has long-existing scientific and practical applications. In recent years, there was an explosive growth in a new generation of intelligent technologies due to significant progress in computing power and mathematics. Artificial Intelligence has the potential to solve many problems in medical diagnostics (decision support, morphometry, workflow automation, quality control, etc.). Before routine use, AI-based systems must undergo clinical trials to evaluate the diagnostic accuracy and obtain a status of the medical device.

Although legislative nuances may vary, in the Russian Federation medical device is defined per the Federal Law No. 323 approved on November 21, 2011. A medical device is any tool, device, equipment, materials and other products used for medicinal purposes (separately or in combination with each other), as well as accessories necessary for the use of the products mentioned before, including specialized software. Medical devices are designed for prevention, diagnosis, treatment, and medical rehabilitation, patient monitoring, medical studies, restoring, replacing, changing anatomy or physiology, preventing, or terminating a pregnancy. The functionality of a medical device is not based on pharmacological, immunological, genetic, or metabolic effects on the human body.

Any software that influences the doctor's decision-making, or provides clinically valuable information, carries a potential risk and thereby may cause harm to patients' health. Therefore, such software, including AI-based solutions, is subject to registration as a medical device. Testing clinical effectiveness is an obligatory stage before submitting such software for registration as a medical device. These methodological guidelines form a unique approach for assessing the diagnostic efficacy of software with clinical tests.

Moreover, an AI-based software can be used both independently or complementary to other medical devices.

# THE MAIN PART

## BASIC FRAMEWORK

In diagnostics, Artificial Intelligence is a multidisciplinary area that combines medicine, biology, mathematics, and computer science and models the individual components of doctors' intellectual activities (Table 1).

**Table 1.** AI in radiology: application areas.

| Primary tasks | AI use case | Standard medical task |
|---|---|---|
| Object localization | Detection | Study prioritization, decision support, quality control |
| 1. Object's presence 2. Object type | Classification | Screening, mass preventive examinations Determining lesion's nature |
| Object size and morphology | Segmentation, volumetry, radiomics | Determining the region of interest for segmentation or volumetry Medical morphometry (lesion follow-up, automatic generation of the image descriptors) |
| Speech and text recognition | Natural language processing | Logging, quality control |
| Evaluation of object's changes over time | Detection, classification, segmentation | Evaluation of the pathological changes during follow-up |

After an AI-based algorithm has successfully passed clinical tests, medical staff should strictly regulate its use as a decision support system. Medical professionals may use the results generated by the algorithm. Artificial intelligence is aimed at improving the efficiency of the health care system, staff productivity, reducing risks and errors, standardizing diagnostic results.

The AI-based software should undergo clinical acceptance before registration as a medical device per Article 36.1 of the Federal Law No. 323. The acceptance tests determine whether the software's accuracy and safety comply with the specifications stated by developers.

According to Article 38 of the Federal Law No. 323, a medical device is any tool, device, equipment, materials and other products used for medicinal purposes (separately or in combination with each other), as well as accessories necessary for the use of the products mentioned before, including a specialized software. Medical devices are designed for prevention, diagnosis, treatment, and medical rehabilitation, patient monitoring, medical studies, restoring, replacing, changing anatomy or physiology, preventing, or terminating a pregnancy. The functionality of a medical device is not based on pharmacological, immunological, genetic, or metabolic effects on the human body. Medical products can be considered interchangeable if they are comparable in functionality, quality, and technical characteristics, and can replace each other.

Per the Letter of the Federal Service for Supervision in Health Care, a state marketing authorization is obligatory for medical devices and software intended for:

- controlling and monitoring equipment usage;
- receiving, storing and processing diagnostic data in an automatic mode;
- monitoring human body functions and transmitting the data (including wireless technologies);
- calculating dosage parameters (radiation, drugs, contrast medium, etc.);
- processing and transferring the data from the diagnostic medical equipment to the therapy systems;
- processing medical images (including quality, color, resolution modification, etc.);
- three-dimensional modeling (dentistry, prosthetics, orthopedics, implantology, organ, and bone prototyping, etc. [2]);
- communication between the diagnostic and therapeutic equipment;
- processing digital images.

Also, it should be noted that the "software" refers to medical devices in the following cases [2]:

- processing, storing and transferring medical records;
- providing medical care via telemedicine;
- installation as part of an information system for home care;
- data processing for the electronic registration of medical equipment, medical devices used in hospitals;
- ensuring communication between patients and medical specialists, so that a patient can follow a treatment per the rules defined when using the application, and for population monitoring;
- installation as a part of a blood bank information system, single station or decentralized network.

In the future, randomized studies are required to evaluate the long-term effects of the AI-based software on the quality of medical care, patient's life expectancy and quality of life. The results of such studies can change the current understanding of the benefits and harms of AI. They will also allow to change the recommendations for the practical use of AI-based software. The impact of AI on the psychology and medical decision-making requires particular research. Before scientific evidence is obtained, caution should be exercised as if the harms of AI could outweigh the benefits. Patient safety is an absolute priority.

Thus, AI-based algorithms and software used in radiology belong to medical devices and must undergo clinical trials. The software can be used independently or in combination with other products (both hardware and software, for example, URIS, PACS, RIS, etc.), irrespective of the hardware platform used, as well as the ways of software deployment and providing access to it.

The following clinical test methodology (in the form of preliminary clinical and technical tests) can also be used in pilot projects, research works, and clinical acceptance tests.

# CLINICAL TESTS

The **purpose of clinical tests** is to confirm the effectiveness, safety of use, and compliance of medical device characteristics with the intended use specified by the manufacturer.

As per para. 8 of Article 38 of the Federal Law of November 21, 2011 No. 323-FZ "On the Fundamentals of Health Protection in the Russian Federation", for the purpose of state registration of medical devices, in the order established by Roszdravnadzor, conformity assessment is carried out in the form of technical tests, clinical tests, and examination of the quality, effectiveness, and safety of medical devices.

As per para. 4 of Registration Rules approved by Regulation No. 1416, **technical tests** are conducted to determine the compliance of the characteristics of the medical device with regulatory documentation, technical and operational documentation of the manufacturer, and to provide further decision on the possibility to conduct clinical tests. It is the manufacturer who determines which regulatory documents its product complies with.

**Clinical tests** are a developed and planned systematic study undertaken to evaluate safety and effectiveness of the medical device. Clinical tests may only be conducted upon successful completion of technical tests.

**For the admission criteria of AI-based software to clinical tests** (list and questionnaire), see Annex 1 and 2. The weight and significance of each criterion can be determined individually for each situation. The criteria form a technical specification, creating a set of requirements for the AI-based software. Also, developers should state *application purposes and clinical scenarios* (Table 1) for a particular modality, use case, and workflow.

Clinical tests are organized per current legislation and methodology for assessing the quality, effectiveness, and safety of medical devices. AI-based software should be evaluated on the *clinical relationship (validity) between selected data and algorithm (concept, measurement, conclusion) per the intended purpose* [2].

The ***clinical evaluation*** of software that is based on artificial intelligence and intended for use in radiology is carried out in two stages (Figure 1):

1. Analytical validation.
2. Clinical acceptance.

*Analytical validation* refers to the evaluation of the correctness of input data processing by the software to create reliable output data, which is performed using labeled reference datasets. Analytical validation (in terms of clinical tests) of the AI-based software includes 6 stages (Table 2).

**Table 2.** Stages of analytical validation of the AI-based software intended for use in radiology.

| Stage | Goal | Brief description | Deliverables |
|---|---|---|---|
| 1 | 2 | 3 | 4 |
| I. Questionnaire | (1) collection of information for the service rating (goals, certification, evidence, functionality, contract); (2) evaluation of the clinical problem to be solved by the service and its applicability in the workflow of radiology departments. | The legal entity that is the AI service developer is requested to fill in a questionnaire (based on Annex 2; for an example, see Annex 3); then, the responses in the questionnaire are analyzed by the commission. | Completed questionnaire |
| II. Self-test | Technical readiness check (seamless integration with the health care information system, processing data from all types of diagnostic devices, providing quality results to health care professionals) | The developer is provided with a sample dataset (usually, 5 to 10 studies) for self-assessment[5] | Files with the results of dataset processing by the software |
| III. Interview | (1) detailed description of the clinical problem and application scenario; (2) evaluation of the legal entity for compliance with the culture of quality and organizational excellence (CQOE). | The company's representatives are interviewed according to the protocol (Annex 4 and 5). | Answers to questions in the protocol |
| IV. Online test | Evaluation of the service performance in real time on a limited dataset that includes a sample of illustrative cases | Data analysis and interpretation of results are carried out with the direct involvement of the commission | Report on the AI algorithm/service performance for each study |

[5] Sample datasets for self-testing are available at ai.npcmr.ru.

| V. Evidence test | Retrospective verification of the claimed accuracy and other parameters of the algorithm performance on the reference dataset. | The software is tested on the reference dataset prepared in accordance with the clinical problem solved by the software. | Evaluation of diagnostic accuracy (analytical validation – see below). |
| --- | --- | --- | --- |
| VI. Final evaluation | (1) Integrated assessment of the software performance and readiness for test operation in real clinical settings of the radiology department; (2) Integrated assessment of the legal entity that is the software developer. | The commission analyzes all the test stages and assigns preliminary scores to the software and legal entity. | Decision on the admission of the service software to integration with the current medical information system (RIS/ PACS) to evaluate the software performance in routine workflow of radiology departments |

*Clinical acceptance (evaluation of the performance by using the software within a standard operating process)* consists of two components:

- clinical correlation (evaluation of whether there is a reliable clinical relationship between the results and the target clinical condition);

- clinical validation (confirmation of achievement of the intended goal for the target population in the clinical workflow through the use of accurate and reliable output data).

The first stage is done by:

a) evaluating the reliability of the legal entity (software developer), including by checking for compliance with the culture of quality and organizational excellence;

b) generating and analyzing diagnostic data for the intended use of software:

- a prepared reference biomedical dataset is created;

- an automated analysis of the dataset is performed, imitating a diagnostic process;

- the results are compared to the prepared reference dataset;

- a mathematical and statistical analysis is performed.

The second stage is software deployment in a routine workflow. Procedures performed with the help of AI-based algorithm are evaluated for performance and quality. That includes multiple timing studies involving several medical specialists at the varying day times and shifts and retrospective analysis of the AI-based algorithm results. The possible risks of the second stage should be minimized, providing safety for patients and their legal representatives.

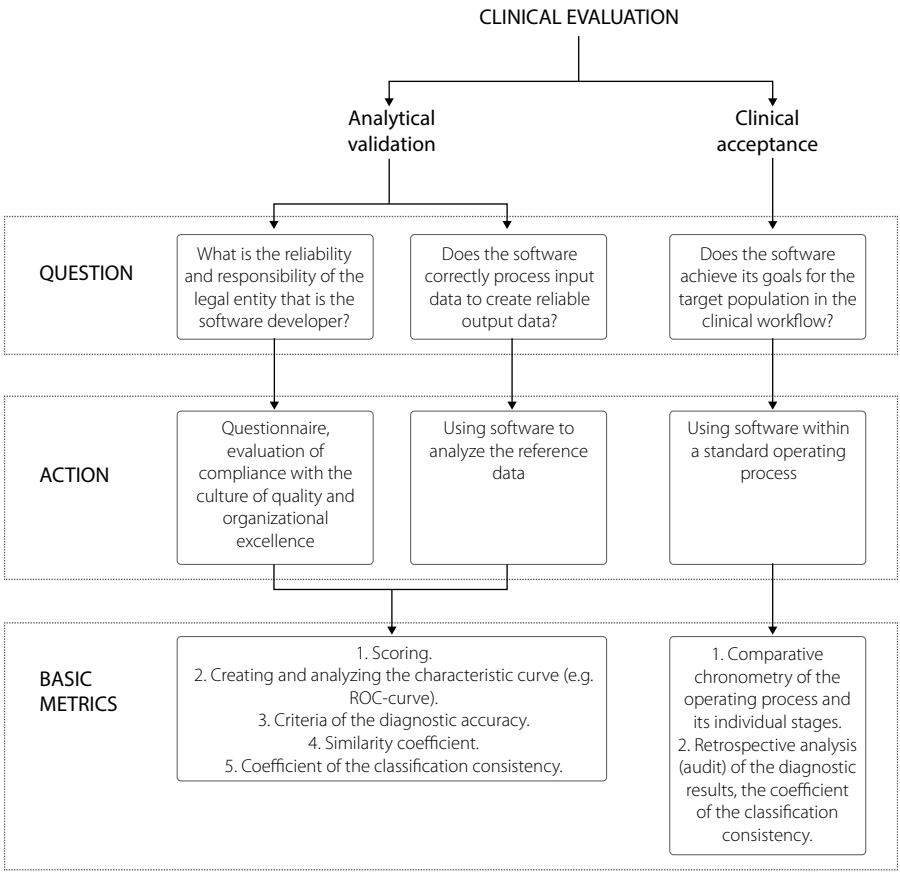

*Figure 1* – Flowchart for clinical evaluation of the AI-based software in radiology.

A protocol of clinical acceptance should include:
- details on the dataset (reference biomedical data), including the state registration certificate (advisable), data type/volume, tagging information;

- information about the software installation process in the medical center;
- information on the use of computing power provided by the developer;
- information on the uninstallation process after the test completion (since the software is installed and used on the equipment before receiving a registration certificate).
-

In general, the software should be predesigned according to the checklist (Annex 6).

## *Preliminary clinical and technical tests*

Preliminary clinical and technical tests (PCTT) are a designed and planned study for preliminary, scientific and practical evaluation of the safety and effectiveness of the AI-based software.

PCTT are conducted under a simplified and adapted clinical test methodology to gain additional knowledge about the performance and limitations of the AI-based software. PCTT do not represent a substitution or equivalent to clinical tests.

PCTT can include stages of analytical validation and clinical acceptance, or consist of analytical validation alone (with the mandatory implementation of 6 stages descried in Table 2). Based on the PCTT results, a report is issued (Annex 7).

PCTT are applicable in the context of research and development and as a tool to evaluate medical and technical readiness of the AI-based software for further in-depth study in real clinical settings.

# QUALITY METRICS

## *Diagnostic accuracy assessment*

An analytical stage consists of designing the diagnostic study (Annex 6) and is assessed with the standard metrics.

The index test is an automated analysis of medical data by the AI-based software.

The biomedical dataset tagging, which was performed according to the methodology, is used as a reference test.

A confusion matrix is built to compare the index and reference tests. It contains the absolute values for each comparison (Table 3). Then, the developers should select and calculate the relevant metrics (Table 4) with the reliability within 95% of the confidence interval.

**Table 3.** AI-based software in radiology: results combined to the confusion matrix (as shown for a response that is a binary variable[6]).

| Result type | | | |
|---|---|---|---|
| True-positive | True-negative | False-positive | False-negative |
| TP | TN | FP | FN |
| The software revealed pathology when it is present | The software did not reveal pathology in its absence | The software revealed pathology in its absence | The software did not reveal pathology when it is present |

**Table 4.** Typical metrics for the analytical evaluation[7].

| AI use case | Basic metrics |
|---|---|
| Detection | The standard set of metrics |
| Classification | The standard set of metrics |
| Segmentation | Similarity coefficient |
| Natural language processing | Classification consistency ratio |

Various combinations of metrics can be used depending on the problem to be solved, for instance, ROC analysis and classification consistency ratio. Basic indicators, such as sensitivity, specificity, AUC should be used to define each type of artificial intelligence result.

---

[6] This type of response corresponds to most typical tasks; however, most other tasks can almost always be reduced to a binary classification task.

[7] These metrics are basic and mandatory for the evaluation. In addition, other metrics can be used as applicable to a specific task.

## Standard set of metrics

Standard metrics are used to compare the diagnostic performance of index test relative to the reference test (Table 5).

**Table 5.** The standard set of diagnostic metrics.

| Indicator | Value | Formula |
|---|---|---|
| 1 | 2 | 3 |
| Sensitivity | The probability that the index test is positive in the presence of pathology | TP / (TP + FN) |
| Specificity | The probability that the index test is negative in the absence of pathology | TN / (TN + FP) |
| Accuracy (general validity) | Level of correspondence (the index test to the reference test) | (TP + TN) / (TP + TN + FP + FN) |
| The likelihood ratio of a positive result | The mathematical relation between the probability of presence and absence of the target pathology with positive index test | sensitivity / (1-specificity) |
| The likelihood ratio of a negative result | The mathematical relation between the probability of presence and absence of the target pathology with negative index test | (1-sensitivity) / specificity |
| Positive predictive value | Likelihood of pathology with positive index test | TP / (TP + FP) |
| Negative predictive value | Likelihood of the absence of pathology with negative index test | TN / (TN + FN) |
| False positive rate | False positive probability | 1–specificity |

All metrics, except the likelihoods, are **evaluated** in the range of 0-1 or in percentage from 0 to 100%:

| Evaluation |
|---|
| <0.6 – unsuitable |
| 0.61 - 0.8 – revision required |
| >=0.81 – admissible for clinical validation |

The likelihood ratio of a positive result should be as large as possible whereas the likelihood ratio of a negative result should be as small as possible.

## Characteristic curve (ROC)

A receiver operating characteristic curve is a graphical plot illustrating the diagnostic ability of a binary classifier system as its discrimination threshold is varied. The ROC curve is created by plotting the true positive rate against the false positive rate. It is a metric of diagnostic value: area under the curve (AUC) – area bounded by ROC-curve and horizontal coordinate.

**Classic ROC-curve**: rpa diagram that illustrates the dependence of sensitivity on 1-specificity (1-specificity on the horizontal coordinate and sensitivity on the vertical coordinate).

When analyzing the ROC-curve, it is necessary to determine the "cut-off" value. It can be done by various methods, including but not limited to the following:

1) minimum distance from the upper left corner to the ROC-curve (d minimum);

2) Youden index, revealing the maximum distance from the diagonal line to the ROC-curve (Figure 2).

The "cut-off" should be defined per the research objectives.

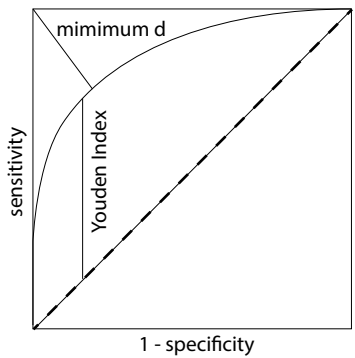

*Figure 2* – Determining the optimal cut-off value.

The area under the curve (AUC) is **evaluated** in the range of 0-1:

| Evaluation |
|---|
| <0.6 – unsuitable |
| 0.61 - 0.8 – revision required |
| >=0.81 – admissible for clinical validation |

## Classification consistency ratio

The consistency (concordance) is a mathematical definition of the agreement between two experts regarding a single phenomenon.

The formula, according to Cohen's method (Cohen's kappa) (Cohen J., 1960):

$$K = (P_0 - P_e) / (1 - P_e),$$

where $P_0$ is the proportion of cases where the measurements coincided, and $P_e$ is the expected proportion of cases with random coincidence. Calculations are as follows:

| | Expert #1 | | | |
|---|---|---|---|---|
| | | 1 | 2 | Total |
| **Expert #2** | 1 | $P_{11}$ | $P_{12}$ | $P_{1\_}$ |
| | 2 | $P_{21}$ | $P_{22}$ | $P_{2\_}$ |
| | Total | $P_{\_1}$ | $P_{\_2}$ | 1 |

$$P_0 = P_{11} + P_{22}$$

$$P_e = P_{\_1}P_{1\_} + P_{\_2}P_{2\_}$$

It is **evaluated** in the range of 0-1 or in percentage from 0 to 100%:

| Evaluation |
|---|
| <0.6 – unsuitable |
| 0.61 - 0.8 – revision required |
| >=0.81 – admissible for clinical validation |

## Similarity coefficient

Binary indicator to quantify the similarity degree of biological objects. The formula, according to Dice-Sörensen's method (Dice L., 1945 Sörensen T., 1948):

$$DSC = 2|A \cap B| / (|A| + |B|),$$

где в числителе – удвоенное количество совпавших оценок, в знаменателе - общее количество оценок.

For sets with the same power, it is **evaluated** in the range of 0-1:

| Evaluation |
|---|
| <0.6 – unsuitable |
| 0.61 - 0.8 – revision required |
| >=0.81 – admissible for clinical validation |

*Clinical acceptance*

*Timing study* is a method of evaluating the software's effectiveness by recording and measuring the duration of actions performed within a workflow. For example, the report turnaround time of a screening chest radiography with and without an AI-based solution.

An *audit* is a retrospective analysis with the goal of quality assurance. In radiology, the audit of AI-based algorithm is performed to determine the agreement with the recommended standards [4]. The audit allows assessing the quality of medical aid, including the diagnostics with and without AI; the significance of differences is determined statistically.

# REFERENCE DATASET REQUIREMENTS

For the analytical stage of clinical and technical or clinical tests, reference data is needed, i.e. biomedical datasets.

Dataset is a set of data that has been pre-prepared (processed) as per the legislation of the Russian Federation on information, information technologies, and information protection, and is necessary for the development of software based on artificial intelligence[8]. In medicine and health care, reference dataset is a structured set of diagnostic data, including diagnostic images and information on pathological changes on images; structured cases of medical care and related electronic medical documents from electronic medical records; libraries of keywords, phrases and their critical combinations. If the dataset contains confirmed information on the final diagnosis for each case, then it is called "verified".

The reference dataset is used for the analytical validation of an AI-based software in radiology.

The size of the reference dataset (i.e., the sample size) to evaluate the diagnostic performance of the AI algorithm is estimated by the sample proportion (p) calculation methods.

For each case (clinical scenario), the desired diagnostic accuracy parameters to evaluate the AI algorithm should be specified; expected proportion with the desired precision of estimate (width of the confidence interval) should be specified. It should be noted that the higher the desired precision of estimate is (i.e., the narrower the confidence interval is) and the closer the proportion is to 50%, the greater number of labeled studies will have to be included in the dataset.

A specialized medical research organization should audit the reference dataset.

We recommend using reference datasets with the state registration for clinical tests.

The reference dataset should contain the following information:

1) state registration certificate number;

2) population characteristics (gender and age, race, geographic, etc.); depersonalization data; medical centers sourcing the dataset formation;

3) imaging study characteristics (anatomical region(s), modality, diagnostic device, study protocol);

4) target pathology per the International Classification of Diseases;

---

[8] National Strategy for the Artificial Intelligence Development for the period up to 2030 (approved by the Decree of the President of the Russian Federation of October 10, 2019 No. 490 "On the Development of Artificial Intelligence in the Russian Federation"

5) the total number of clinical cases, studies, images, reports and their distribution in diagnostic groups;

6) the normal-to-abnormal ratio ("pathology" cases can be divided into several subcategories);

7) verification data (histopathological or other "final" diagnosis);

8) tagging methodology (scientific publications, guidelines, or patents).

The reference dataset should meet the following requirements:

1) the normal-to-abnormal ratio should reflect the prevalence of the target pathology in the population;

2) several medical centers should source the reference dataset to introduce the data heterogeneity;

3) demographic, socio-economic characteristics and basic health indicators in the reference dataset should correspond to the population's average characteristics in the target region;

4) the proposed size of the reference dataset should be justified per statistical considerations, and the desired diagnostic accuracy by the main metrics indicated above;

5) reference datasets used in clinical tests for registering the software as a medical device should not be publicly available (to exclude the possibility of training AI algorithms on reference datasets).

# RESULT REGISTRATION

In the evaluation of clinical and technical or clinical tests, the following should be considered:

- the software is tested with the dataset and on all necessary equipment (including other medical devices) designed to ensure the proper functioning of the AI-based software;

- clinical tests should utilize all software modules and functions per the developer's specification;

- all items of evidence-based software should be assessed (clinical data, interchangeable medical devices, etc.), as submitted by the developer to the medical testing organization;

- the medical organization conducting the clinical tests must evaluate the software's documentation, including technical and operational documentation, accessibility/clarity to the relevant specialists, as the contents must ensure a proper, effective and safe intended use.

The results for any medical device should be registered. In Russian Federation, registration is carried out per the Order of the Ministry of Health dated January 9, 2014 No. 2n "On the Approval of Assessing the Conformity of Medical Devices in the Form of Technical Tests, Toxicological Studies, Clinical Tests for the State Registration of Medical Devices".

We recommend creating the documentation with the detailed report of data evaluation and analysis per the updated STARD 2015 checklist (Annex 6).

# CONCLUSION

Some algorithms and tools based on intelligent technologies have already been successfully used in medical imaging/radiology. Further development of artificial intelligence should be targeted at solving specific problems.

The implementation of these recommendations allows for unifying clinical acceptance of an AI-based software in radiology. Such tests are a mandatory component for the registration of as a medical device.

# Criteria for the admission of AI-based software to clinical tests (preliminary clinical and technical tests)

**1.** Goals:

- The software provides a preliminary automatic analysis of medical images to improve the quality and speed of the radiology workflow;
- The software ensures a prioritization in the worklist according to the automatically revealed pathology;
- The software provides a preliminary comparative analysis of studies of a single patient at different time points (dynamic study);
- The software provides support in medical decisions;
- The software automatically prepares a draft of the radiology report based on the results of the analysis.

**2.** Certification:

- The medical device has passed technical tests in an accredited laboratory;
- Approvals of FDA and/or CE certification (class II); actual implementations of the currently working software in medical centers: at least 2 independent institutions; more than 6 months of operation; at least 1000 successfully completed studies (confirmed by radiologists) for each task (if the software solves several tasks);
- Scientific articles (original research works) published in peer-reviewed journals indexed by "Scopus" and/or "Web of Science" and included in the first and second quartile according to the "International Scientific Journal & Country Ranking";
- Proven diagnostic accuracy AUC>=0.81 (classic ROC curve) and increase of the radiology workflow efficiency (based on the comparison of reporting speed with and without the software, including timing).

**3.** Security:

- Compliance with the requirements of the legislation of the Russian Federation in the field of personal data, information security and health protection;
- Availability or readiness to deploy server capacities necessary for software operation within the Russian Federation.

**4.** Evidence:

- Once the development was completed, the accuracy of algorithms was assessed on independent data[9];

---

[9] The medical dataset for testing was different from the dataset used for training, calibration and validation of the algorithm (that is, testing was performed on data that the algorithm had not previously encountered).

- Diagnostic accuracy tested on population whose characteristics are similar to those where the AI is intended to be used;
- Annual update of diagnostic accuracy information.

**5.** Standardization:
- Automated analysis of diagnostic images in DICOM standard;
- Support of the HealthLevel7 (HL7) / FHIR standard (in particular, the system should provide an exchange of messages on the completion of automatic image analysis, pathology detection and classification);
- Use of recommended classifications (RADS, MAGNIMS, etc.) when making reports.

**6.** Integration:
- Availability or readiness to develop means for "seamless" integration with information systems working in the field of health care of the given subject of the Russian Federation, medical information systems;
- Availability of tools for integration with PACS and RIS (DICOM "query and retrieve");
- Possibility of "seamless" integration with PACS / RIS, provided only by software.

**7.** Functionality:
- Ability of stream processing and subsequent sending of series to PACS, extended with the results of the AI analysis;
- Possibility to combine series of native images and those containing the results of the AI analysis;
- Identification of findings (nosologies, pathological signs, deviations from norm) for a given modality or working process according to the technical specifications;
- Classification (determination of type and variety) of findings, including by ICD-10 if required by the technical specifications;
- Providing information on the likelihood of target pathology;
- Automatic preparation and uploading of a draft of findings description to PACS (subject to the structure of description template preinstalled in PACS);
- Automatic analysis result displayed by standard PACS tools, including DICOM graphics capabilities;
- Availability of description protocol templates with their automatic generation and sending to RIS / MIS via HL7 / FHIR messages;
- Availability of a built-in accuracy assessment tool;
- Maximum processing time of a single radiology study does not exceed

the set time period (which is defined individually based on the clinical scenario, infrastructure characteristics, etc.; for example, 60 seconds without considering the time for data transfer). When evaluating the dynamics (comparing the studies), the analysis may take more than 60 seconds, but not more than 60 seconds for one study;

- Control function ensuring the start of the automated analysis only after receiving the "status complete by radiographer" HL7 message from RIS;

- HL7 software reports on the completion of automated analysis and pathology detection for the prioritization of studies in the worklist;

- Possibility of automated search for similar studies for comparison in the PACS / RIS database;

- For each analyzed study, the software generates a declaration containing the software (algorithm) name, a list of the revealed findings (nosologies, pathological signs, deviations from norm) with indication of sensitivity and specificity for each type; the declaration is formed in DICOM SR, Overlay etc., and transmitted to PACS as a separate series;

- The result of software operation is series of images (DICOM format), with:

a) a number of slices similar to those in the original series for a simultaneous viewing by radiologist;

b) information on each slice contains the software name, version, diagnostic accuracy, the verification date and the exact time of completed study;

c) possibility to provide additional series with the analysis results (e.g. summary tables with the revealed findings in dynamics and/or particular images of findings).

**8.** Contract:

- The legal entity that is the software developer must have a quality management system;

- The legal entity that is the software developer must have a version preparation and control policies;

- Regular system updates, including those for diagnostic accuracy information;

- Software updates included in the price;

- All medical data, related materials and software results are the property of the customer.

Annex 2

# Questionnaire for the admission of software based on artificial intelligence / computer vision to a preliminary test operation (preliminary clinical and technical tests)

| Section | Metrics | Answer (1-yes / 0-no) | Comments |
|---------|---------|----------------------|----------|
| 1. Goals | 1.1. The software provides a preliminary automatic analysis of medical images (DICOM files) to improve the quality and speed of the radiology workflow.<br>1.2. The software ensures a prioritization in the worklist according to the automatically revealed pathology.<br>1.3. The software automatically prepares a draft of the radiology report based on the results of the analysis.<br>1.4. The software provides a preliminary comparative analysis of studies of a single patient at different time points (dynamic study). | | |
| 2. Certification | 2.1. Approvals of FDA and/or CE certification (class II).<br>If the answer to clause 2.1 is "no", there should be positive answers to clauses 2.2 and 2.3.<br>2.2. Actual implementations of the currently working software in medical centers:<br>- at least 2 independent institutions;<br>- more than 6 months of operation;<br>- at least 1000 successfully completed studies (confirmed by radiologists) for each task (if the software solves several tasks).<br>2.3. Scientific articles (original research works) published in peer-reviewed journals indexed by "Scopus" and/or "Web of Science" and included in the first and second quartile according to | | |
| | the "International Scientific Journal & Country Ranking"; proven diagnostic accuracy AUC>=0.81 (classic ROC curve) and increase of the radiology workflow efficiency (based on the comparison of reporting speed with and without the software, including timing). | | |
| 3. Evidence | 3.1. Once the development was completed, the accuracy of algorithms was assessed on independent data, i.e. biomedical dataset for testing differed from the one used for training, development and validation. That is, clinical tests were performed on data unknown to the algorithms.<br>3.2. Diagnostic accuracy was tested on the population whose characteristics are similar to those where the AI is intended to be used.<br>3.3. Annual update of diagnostic accuracy information. | | |
| 4. Functionality | 4.1. Availability of a built-in accuracy assessment tool (if applicable).<br>4.2. Maximum processing time of a single radiology study does not exceed the set time period (which is defined individually based on the clinical scenario, infrastructure characteristics, etc.; for example, 60 seconds without considering the time for data transfer). When evaluating the dynamics (comparing the studies), the analysis may take more than 60 seconds, but not more than 60 seconds for one study.<br>4.3. The result of software operation is series of images (DICOM format), with:<br> - a number of slices similar to those in the original series for a simultaneous viewing by radiologist;<br> - information on each slice contains the software name, version, diagnostic accuracy, the verification date and the exact time of completed study; | | |
| | - possibility to provide additional series with the analysis results (e.g. summary tables with the revealed findings in dynamics and/or particular images of findings). | | |
| 5. Contract | 5.1. The legal entity that is the software developer must have a quality management system.<br>5.2. Regular system updates, including those for diagnostic accuracy information.<br>5.3. Software updates included in the price.<br>5.4. All medical data, related materials and software results are the property of the customer. | | |

**Annex 3**

# Research and Practical Clinical Center for Diagnostics and Telemedicine Technologies: Questionnaire about the software based on AI technologies/computer vision

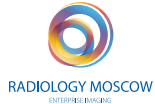

RADIOLOGY MOSCOW
INTERNET IMAGING

## QUESTIONNAIRE ABOUT THE SOFTWARE BASED ON AI TECHNOLOGIES/COMPUTER VISION

| Section | Metrics | Answer | Comments, clarifications, suggestions |
|---|---|---|---|
| Company name | | | |
| 1. Solution | 1.1. Solution name. | Free response in the «Comments, clarifications, suggestions» field | |
| | 1.2. Imaging modality. | Free response in the "Comments, clarifications, suggestions" field | |
| | 1.3. Anatomical area. | Free response in the "Comments, clarifications, suggestions" field | |
| | 1.4. Application area. | ☐ Oncology ☐ Pulmonology ☐ Cardiology ☐ Neurology ☐ Chronic diseases ☐ Medical emergencies | |
| | 1.5. Age group. | Free response in the "Comments, clarifications, suggestions" field | |
| 2. Clinical application scenario | 2.1. Type of medical care. | ☐ Planned ☐ Emergency | |
| | 2.2. Level of medical care. | ☐ Primary ☐ Secondary ☐ Tertiary | |
| | 2.3. Stage of medical care. | ☐ Prehospital ☐ Hospital ☐ Outpatient | |
| | 2.4. Stage of the clinical workflow, at which the results are provided to the user. | Notification: ☐ Before viewing the worklist ☐ In the worklist ☐ After a study is opened ☐ Upon request (second opinion) ☐ Audit of a completed report | |
| | 2.5. Ways in which the results are provided. | ☐ Binary assessment of a pathology presence ☐ Score/categorical assessment of a pathology presence ☐ Free text description ☐ Study report template ☐ Recommendations for patient management ☐ Visualization of the findings on images ☐ Schematic view | |
| | 2.6. Decisions taken by the users based on the obtained results. | ☐ No decisions ☐ Selection of further methods to analyze the study ☐ Referring to a specialist ☐ Additional examination with other diagnostics methods ☐ Additional examination with the same method after a certain time period | |
| | 2.7. Practical goals. | ☐ Time reduction between the diagnostic study finish and the report completion ☐ Improvement of the quality of work (e.g., reduction of the clinically significant errors) ☐ Audit of radiologists performance | |
| 3. Risks | 3.1. What class of medical software does the proposed AI service belong to? (Only one answer is possible)<br><br>*Software classification scheme is given at the end of the questionnaire.* | Class 1<br>Class 2a<br>Class 2b<br>Class 3 | |
| 4. User categories | 4.1. Who can use the AI service? | Specially trained healthcare professionals<br>Patient supervised by specially trained healthcare professionals<br>Patient without supervision of specially trained healthcare professionals | |
| 5. Functional capabilities | 5.1. Automated analysis of medical images (DICOM files). | Yes<br>No | |
| | 5.2. Prioritization in the worklist according to the automatically revealed pathology. | Yes<br>No | |
| | 5.3. Automated preparation of a draft radiology report based on the results of the analysis. | Yes<br>No | |
| | 5.4. Preliminary comparative analysis of studies of a single patient at different time points (dynamic study). | Yes<br>No | |
| 6. Certification | 6.1. Approvals of FDA and/or CE certification (class II). | Yes<br>No<br>In progress | |
| | 6.2. Actual implementations of the currently working software in medical centers:<br><br>– at least 2 independent institutions;<br>– more than 6 months of operation;<br>– at least 1000 successfully completed studies (confirmed by users) for each task (if the software solves several tasks).<br><br>*Please, provide a list (including contact information) of hospitals/clinical centers where your product was used or is being used and copies of the contracts/agreements with these sites (without confidential details) and a detailed report on the implementation results.* | Yes<br>No<br>In progress | |
| | 6.3. Scientific articles (original research works) published in peer-reviewed journals indexed by Scopus and/or Web of Science and included in the first and second quartile according to the International Scientific Journal & Country Ranking; proven diagnostic accuracy AUC≥0.8 (classic ROC curve) and increase of the radiology workflow efficiency (based on the comparison of reporting speed with and without the software, including timing). | Yes<br>No<br>In progress | |
| 7. Evidence | 7.1. Once the development was completed, the accuracy of algorithms was assessed on independent data, i.e. medical database for testing differed from the one used for training, development and validation. That is, clinical tests were performed on data unknown to the algorithms. | Yes<br>No | |
| | 7.2. Diagnostic accuracy was tested on data that included Caucasoid and Mongoloid races (if the information is not available, on data of Russian citizens) | Yes<br>No | |
| | 7.3. Periodic update of diagnostic accuracy information (please, specify the period in comments) | Yes<br>No | |
| | 7.4. A list of criteria that you used to evaluate the achievement of the practical goal (item 2.7). *Please, send summaries or reference letters written by the product users as well as the documents confirming the achievement of the claimed practical goal(s).* | Free response in the «Comments, clarifications, suggestions» field | |
| **8. Functionality** | 8.1. Availability of a built-in accuracy assessment tool. | Yes No N/A | |
| | 8.2. Processing time of a single radiology study (specify in sec.) *Please, specify system requirements.* | sec | |
| | 8.3. The result of software operation is series of images (DICOM format), with: – the possibility to synchronize with the original images of the study; – information on each slice contains the software name, version, diagnostic accuracy, the verification date and the exact time of completed study; – possibility to provide additional series with the analysis results (e.g. summary tables with the revealed findings in dynamics and/or particular images of findings). *Please, provide a list and examples of output files that your service delivers in clinical practice as well as the workstation screenshots to demonstrate the result presentation format.* | Yes No | |
| **9. Contract** | 9.1. Regular system updates, including those for diagnostic accuracy information. | Yes No | |
| | 9.2. Software updates included in the price. | Yes No | |
| | 9.3. All medical data, related materials and software results are the property of the customer. | Yes No | |
| **Completed by (full name and contact details)** | | | |
| **Other information (if applicable)** | | | |

**Software Classification Scheme**

**When identifying the software class, please specify the categories applicable to the service**

When classifying the software that is a medical device, only one class may be assigned to each software (Table 1):

**Class 1: Low-risk software**
**Class 2a: Medium-low risk software**
**Class 2b: Medium-high risk software**
**Class 3: High-risk software**

Table 1. Software classification scheme

| Clinical situation category | Information value | | |
|---|---|---|---|
| | I | II | III |
| A | 3 | 2b | 2a |
| B | 2b | 2a | 1 |
| C | 2a | 1 | 1 |

EXPLANATORY NOTES:

**INFORMATION VALUE**

*I – Crucial information*
Information that is (a) crucial to make an informed clinical decision when making a diagnosis and/or providing treatment to a patient, and (b) used to take immediate and timely action:

– When treating, preventing, or alleviating disease manifestations through the use of medicines, medical devices, or other treatment methods;
– To detect diseases (i.e., for diagnosis or screening).

*II – Information that requires clarification*
Information that requires clarification and/or more details due to its insufficiency to make an informed clinical (medical) decision:

– Information on the safe and effective use of medicines and medical devices that is used in the treatment of diseases;
– Information used to predict the risk of disease development, as well as supporting information used to identify the signs and symptoms of the disease or make a preliminary or final diagnosis;
– Classification or identification of early symptoms of the disease.

*III – Information intended to provide the long-term treatment*
Information that (a) is intended to provide the long-term treatment, (b) does not require immediate action, and (c) is meant to inform about diseases:

– Information on the options available to diagnose, treat, prevent, or alleviate disease manifestations;
– Information obtained by the software by collecting relevant data (e.g., data on patient diseases, used medicines or medical devices, etc.).

**CLINICAL SITUATION CATEGORIES**

**Category A**
The clinical situation is classified into Category A if the software is intended for use:

– In case of emergency medical care;
– In severe, extremely severe, and terminal general condition of the patient;
– When determining the need for major therapeutic or surgical intervention;
– In the diagnosis or treatment of diseases that pose a high risk to public health and/or for high-risk patients (including for vulnerable population group).

In this situation, the software can only be used by specially trained healthcare professionals.

**Category B**
The clinical situation is classified into Category B if the software is intended for use:

– In case of urgent medical care;
– In moderate general condition of the patient;
– If the disease or condition does not require major therapeutic intervention;
– In the diagnosis or treatment of diseases that pose a moderate risk to public health.

The software can be used both by specially trained healthcare professionals and by patient or other individual supervised by specially trained healthcare professionals. If the software is used by patient or other individual without supervision of specially trained healthcare professionals, this clinical situation is classified into Category A.

**Category C**
The clinical situation is classified into Category C if the software is intended for use:

– In case of routine medical care;
– In satisfactory general condition of the patient;
– If the disease requires minor therapeutic intervention (usually, non-invasive) or long-term medical supervision;
– In the diagnosis or treatment of diseases that pose a low risk to public health.

*The software can be used both by specially trained healthcare professionals and by patient or other individual without supervision of specially trained healthcare professionals.*

*N.B.: If more than one provision may apply to the software,*
*the software class is determined according to the highest potential risk.*

**Annex 4**

## Interview protocol for evaluating the reliability and quality of the developer of the AI-based software intended for use in radiology

**General requirements:**

- Radiology workstation displaying the results obtained by the AI service algorithms;
- Possibility to load and process studies using the AI service during a face-to-face meeting;
- Availability of AI processing timing from loading to providing the result;
- Information on the architecture of the models used;
- Access to the code;
- Presence of a developer and a big data analyst who can provide consultation on the code and structure of algorithms, training process, and validation of the algorithm(s);
- Presence of a doctor who uses the AI service in his/her practice, or documents confirming successful clinical acceptance, indicating contact details (optional).

### Evaluation of clinical significance

1. What kind of clinical problem does the AI service solve?
2. Is the use of the AI service for the selected clinical problem justified in the context of evidence-based medicine?
3. How does the result provided by the AI service affect patient management strategy?
4. Does the quality of doctor's work improve?
5. Does the speed of doctor's work increase?

### Analytical validation

*A) Technical readiness*

1. Self-test.
2. Timing, measuring the time required to process 1 study.
3. Output data format.
4. Visual representation of the results obtained by the AI service on the radiology workstation.
5. Output data format (DICOM, JPEG, PNG, BMP, AVI (copy)).

*B) Details on creating and applying the AI model*

    1.   What databases were used in building the AI models (from open sources, external partners, in-house)?

    2.   Balancing classes in databases.

    3.   Total amount of data used to build the AI model for each clinical problem.

    4.   Architecture of the AI model.

    5.   Libraries and resources used to build the model.

    6.   Approaches to the model development (from scratch, transfer learning, retraining of existing architecture).

    7.   Type of machine learning (supervised / unsupervised / semi-supervised / weak supervision).

    8.   Validation of the model (independent validation set, cross-validation).

    9.   Time required to create AI models. Available computational resources.

    10. Are the AI models retrained? How is the versioning organized?

    11. Does the AI service process data other than images?

    12. Can the AI service evaluate technical quality of the study?

    13. Is there any image preprocessing?

    14. Can the AI service evaluate the quality of the study interpretation (i.e., control the radiologist's work)?

    15. Values of accuracy metrics of the AI service performance for each clinical problem (AUC, sens, spec, etc.)

    16. Were the AI models compared to existing analogues that solve the same clinical problems?

*C) Details on the datasets and their tagging used for developing AI models*

    1.   Were any criteria applied for selection of a database for training? What kind of criteria (gender, age, diseases, ICD, normal-to-abnormal ratio)?

    2.   Did the training database contain different studies of the same patient (to be trained to recognize dynamics)?

    3.   What was considered ground truth in each clinical scenario in the tagged database?

    4.   Were any quality control criteria applied to the collected database before tagging?

    5.   What is the business process of tagging?

    6.   Were the taggers selected in a particular way? Any criteria?

    7.   Any quality control of input data and tagging results?

    8.   How many people participated in the data tagging (how many people per 1 study; was the same study tagged by one specialist twice)?

    9.   Were the taggers trained before tagging?

    10. Tagging software: in-house or external? Please specify.

11. Approach in case of inconsistency of taggers' opinions.

12. Business process used by the expert confirming the finding:

13. Worklist;

14. Retagging;

15. Digital footprint (a series of expert's comments addressed to the tagger).

16. Was the tagging process automated (CAD for outlining)?

17. Monitoring of the tagging process.

## Clinical validation

*A) Practical application in clinical workflow*

1.  Was the integration with PACS performed in a medical organization? If not, how was the workflow integration performed?

2.  What is the business process used by the radiologist working with the AI service in practice?

3.  Are the AI service operation instructions available for the doctor?

4.  Questions to the doctor: how was he/she trained and whether AI service operation instructions and opportunities for feedback are available?

5.  Availability of study triage.

6.  Possibility to localize the finding using AI.

7.  Possibility to classify the finding.

8.  Possibility to compare studies over time.

9.  Availability of a template to form a report and/or conclusion.

10. At what stage of business process does the radiologist receive the processing results obtained by the AI service?

11. Availability of feedback from the radiologist on the use of AI. What kind of feedback? If available, is it mandatory?

12. What happens if the radiologist and AI have different opinions (overdiagnosis and underdiagnosis)? How is the final decision on the study made?

13. Availability of any metrics to evaluate the use of AI in clinical practice. Examples (e.g., decreasing the number of errors made by radiologists and reducing the time required to interpret one study).

**Annex 5**

## Evaluation of corporate culture components demonstrated by the developer of the AI-based software intended for use in radiology to comply with the culture of quality and organizational excellence (CQOE)[10]

| Item | Description | Points (20 = satisfactory, 15 = non-critical remarks, 5 = critical remarks, 0 = no result) |
|---|---|---|
| A. | Patient safety. Demonstration of excellence in providing a safe patient experience and emphasizing patient safety as a critical factor in all decision-making processes. | |
| B. | Product quality. Demonstration of excellence in the development, testing, and maintenance necessary to deliver SaMD products at the highest level of quality. | |
| C. | Clinical responsibility. Demonstration of excellence in responsibly conducting clinical evaluation and ensuring that patient-centric issues, including labeling and human factors, are appropriately addressed. | |
| D. | Cybersecurity responsibility. Demonstration of excellence in protecting cybersecurity and proactively addressing cybersecurity issues through active engagement with stakeholders and peers. | |
| E. | Proactive culture. Demonstration of excellence in a proactive approach to surveillance, assessment of user needs, and continuous learning. | |
| Final score (maximum 100) | | |

---

[10] Developing a Software Precertification Program: A Working Model.V.1.0 – January 2019. - https://www.fda.gov/media/119722/download.

<div align="right">**Annex 6**</div>

# Checklist for a detailed description of the results of clinical acceptance of software developed for radiology based on intelligent technologies (according to STARD 2015 [7] with addenda)

| Section and topic | No. | Item |
|---|---|---|
| TITLE | | |
| | 1 | Emphasis on assessing the diagnostic accuracy of software based on intelligent technologies designed for radiology |
| ABSTRACT | | |
| | 2 | Structured summary of study design, materials and methods, results, and conclusions |
| INTRODUCTION | | |
| | 3 | Scientific and clinical data, including the intended use and clinical role of the index test |
| | 4 | Study objectives, hypotheses and endpoints |
| METHODS | | |
| Study design | 5 | Prospective or retrospective study (whether data collection was planned before the index test and reference standard were performed) |
| Sampling | 6 | Eligibility criteria (including targeted pathology) |
| | 7 | On what basis participants were identified (such as symptoms, results from previous tests, etc.) |
| | 8 | Where and when the biomedical dataset was formed (incl. key characteristics: state registration, population characteristics, data depersonalization, targeted pathology and diagnostic groups distribution, studies characteristics, verification, labelling methodology) |
| | 9 | Planned sampling size and way of its calculating |
| | 10a | The sampling was formed sequentially, randomly or in other ways. Justification of sampling size |
| Test methods | 10b | Index test, in sufficient detail to allow replication (including information on mathematical models, neural networks, machine learning methods, samplings for training and calibration used in the development process) |
| | 11 | Reference test ("gold standard"), in sufficient detail to allow replication (the reference test is a labeled (reference) dataset) |

| | 12a | Rationale for choosing the reference test (labelling methodology, the whole dataset) |
|---|---|---|
| | 12b | The activation threshold ("cut-off" point) of the index test, including distinguished pre-specified result categories from exploratory. Method of determining the activation threshold: the minimum distance from the upper left corner to the ROC-curve, Youden index (depending on the study objectives) |
| | 13a | The activation threshold (threshold rules decision, "cut-off") of the reference test, including distinguished pre-specified result categories from exploratory. Method of determining the activation threshold: the minimum distance from the upper left corner to the ROC-curve, Youden index (depending on the study objectives) |
| | 13b | Whether critical clinical and laboratory information and reference standard results were available to the performers / analysts of the index test (blind study or not; in the second case the reason should be justified) |
| | 14 | Whether critical clinical and laboratory information and index test results were available to the participants of labelling / analysis of the reference data (blind study or not; in the second case the reason should be justified) |
| Analysis | 15 | Methods for estimating or comparing measures of diagnostic accuracy (metrics) |
| | 16 | How indeterminate index test or reference standard results were handled |
| | 17 | How missing data on the index test and reference standard were handled |
| | 18 | Any analyses of variability in diagnostic accuracy, distinguishing pre-specified from exploratory |
| RESULTS | | |
| Participants | 19 | Patients characteristics: flow of participants, population for analysis, number of patients included in the study, number of prematurely withdrawn patients, number of skipped or unusable data. In addition to the text description it is recommended to provide a CONSORT diagram |
| | 20 | Baseline demographic and clinical characteristics of sampling (when performing the initial survey; all subjects, as well as subjects with results of both tests) |
| | 21a | Distribution of severity of disease in those with the target condition |

| | 21b | Distribution of alternative diagnoses in those without the target condition (according to the International Classification of Diseases) |
|---|---|---|
| | 22 | Any significant differences in the methods of the index test and reference test |
| Test results | 23 | Combined table of index test and reference test results |
| | 24 | Estimates of diagnostic accuracy and their precision (such as 95% confidence intervals) |
| | 25 | Any adverse events from performing the index test or the reference standard |
| DISCUSSION | | |
| | 26 | Study limitations, including sources of potential bias, statistical uncertainty, and generalizability |
| | 27 | Implications for practice, including the intended use and clinical role of the index test |
| OTHER INFORMATION | | |
| | 28 | Registration number and name of registry (any stage) |
| | 29 | Where the full study protocol can be accessed (including biomedical dataset information for training and calibration) |
| | 30 | Sources of funding and other support; role of funders |

**Annex 7**

# Recommended report form for preliminary clinical and technical tests (PCTT) of the AI-based software intended for use in radiology

| No. | Item | Description |
| --- | --- | --- |
| 1 | Institution | Full name of the institution where the PCTT were conducted |
| 2 | Contact details | Contact details of the institution (address, phone, website, e-mail) |
| 3 | Dates of PCTT | Range of dates when the PCTT were conducted |
| 4 | Summary | Structured summary of study design, materials, methods, results, and conclusions |
| 5 | Purpose, objectives, and endpoints of the PCTT | Purpose, objectives, and endpoints of the PCTT |
| 6 | Reference test (reference dataset) | Detailed description of the prepared reference datasets |
| 6.1 | Data type | Data type(s) (medical records, study results, etc.), modalities, and other characteristics |
| 6.2 | Number of clinical cases included | Number of clinical cases included (patients, study results, etc.) |
| 6.3 | Population characteristics | Population characteristics (race, gender, and other characteristics) |
| 6.4 | Dataset and tagging characteristics | Where and when the reference dataset was formed (incl. key characteristics: state registration of the database, data depersonalization, voluntary informed consent, inclusion/exclusion criteria, sources of clinical cases). How the dataset was prepared (labeled) |
| 6.5 | Pathology characteristics in the dataset | Target pathology and diagnostic groups distribution. Verification method and availability of relevant information in the dataset |
| 6.6 | Dataset generation | Sequentially, randomly, or in other ways. Justification of sample size |
| 6.7 | Data sources | Number and location of medical institutions that provided clinical cases included in the PCTT |

| 6.8 | PCTT independence notice | The dataset used for the PCTT should not be used, either in full or in part, for training or calibration of the index test |
|------|--------------------------|-----------------------------------------------------------|
| 7 | Index test (AI-based software) | Information on the use, installation, accessibility, etc. of the index test (tested AI-based software) |
| 8 | PCTT process | A brief general description of the test process; the use of flowcharts and CONSORT diagram is possible |
| 9 | Result table | Combined table of index test and reference test results |
| 10 | Activation threshold | "Cut-off" point for the reference test and index test, including distinguished pre-specified result categories from exploratory |
| 11 | Diagnostic accuracy parameters | Diagnostic accuracy metrics with a 95% confidence interval (sensitivity, specificity, general accuracy, AUC, etc.) |
| 12 | Limitations | Any PCTT limitations, including sources of potential bias, statistical uncertainty, and generalizability. Any significant differences in the methods of the index test and reference test |
| 13 | Conclusions | Brief summary of the results |
| 14 | Sources of PCTT funding | Statement of the source of funding for the works related to the PCTT |
| 15 | Other information | Any additional information |
| 16 | List of researchers | List of employees of the institution who conducted the PCTT (full name, position, academic title/degree) |
| 17 | Date of signing the report | - |
| 18 | Signature of the responsible person | Personal signature of the person responsible for the PCTT, full name |
| 19 | Signature of the head of institution | Personal signature of the head of institution, full name |
| 20 | Seal of the institution | - |